\documentclass[journal]{IEEEtran}
\usepackage{graphicx}
\usepackage{cite}
\usepackage{picinpar}
\usepackage{amsmath}
\usepackage{url}
\usepackage{flushend}
\usepackage[latin1]{inputenc}
\usepackage{colortbl}
\usepackage{soul}
\usepackage{multirow}
\usepackage{pifont}
\usepackage{color}
\usepackage{alltt}
\usepackage[hidelinks]{hyperref}
\usepackage{enumerate}
\usepackage{siunitx}
\usepackage{breakurl}
\usepackage{epstopdf}
\usepackage{pbox}
\usepackage{booktabs}
\usepackage{threeparttable}
\usepackage{tablefootnote}

\begin{document}
\title{	DSO-VSA: a Variable Stiffness Actuator with Decoupled Stiffness and Output Characteristics for Rehabilitation Robotics}

\author{
	Maozeng Zhang, Ke Shi, Huijun Li, Tongshu Chen, Jiejun Yan, and Aiguo Song, ~\IEEEmembership{Senior Member,~IEEE}

\thanks{This work has been submitted to the IEEE for possible publication. Copyright may be transferred without notice, after which this version may no longer be accessible.}
}

\maketitle

\begin{abstract}

Stroke-induced motor impairment often results in substantial loss of upper-limb function, creating a strong demand for rehabilitation robots that enable safe and transparent physical human-robot interaction (pHRI). Variable stiffness actuators are well suited for such applications. However, in most existing designs, stiffness is coupled with the deflection angle, complicating both modeling and control. To address this limitation, this paper presents a variable stiffness actuator featuring decoupled stiffness and output behavior for rehabilitation robotics. The system integrates a variable stiffness mechanism that combines a variable-length lever with a hypocycloidal straight-line mechanism to achieve a linear torque-deflection relationship and continuous stiffness modulation from near zero to theoretically infinite. It also incorporates a differential transmission mechanism based on a planetary gear system that enables dual-motor load sharing. A cascade PI controller is further developed on the basis of the differential configuration, in which the position-loop term jointly regulates stiffness and deflection angle, effectively suppressing stiffness fluctuations and output disturbances. The performance of prototype was experimentally validated through stiffness calibration, stiffness regulation, torque control, decoupled characteristics,  and dual-motor load sharing, indicating the potential for rehabilitation exoskeletons and other pHRI systems.

\end{abstract}

\begin{IEEEkeywords}
Rehabilitation robotics, variable stiffness actuators, physical human-robot interaction, mechanism.
\end{IEEEkeywords}

% \markboth{IEEE TRANSACTIONS ON Industrial Electronics}%
% {}

\definecolor{limegreen}{rgb}{0.2, 0.8, 0.2}
\definecolor{forestgreen}{rgb}{0.13, 0.55, 0.13}
\definecolor{greenhtml}{rgb}{0.0, 0.5, 0.0}

\section{Introduction}

\IEEEPARstart{S}{troke-induced} neuronal damage significantly impairs upper limb motor function, potentially resulting in hemiplegia \cite{thomas2017repetitive}. In recent years, rehabilitation robotics have emerged as efficient and user-friendly tools for promoting recovery \cite{klamroth2014three,zhang2025two}. Since rehabilitation involves direct interaction with individuals affected by motor impairments, actuator safety and transparency are essential for effective physical human-robot interaction (pHRI) \cite{grosu2017Tmech,2003IJRR}.

To enhance the performance in robotic devices, numerous studies have focused on incorporating compliance into actuators\cite{2017IJRR}. A well-established method is the active strategy, which emulates mechanical compliance by integrating sensory feedback with torque/force control\cite{liulin}. However, ensuring its robustness during interaction may be challenging, particularly under conditions such as sensor failure or low sampling frequency\cite{1998wang}. This limitation has motivated the development of passively compliant actuators with inherent safety, including Series Elastic Actuators (SEAs) and Variable Stiffness Actuators (VSAs)\cite{2023PVSA}. By incorporating elastic elements, SEAs offer benefits including low mechanical impedance, reduced reflected inertia, passive energy storage, and enhanced tolerance to impact loads\cite{2022TMECH}. Nevertheless, the fixed stiffness profiles of these actuators create a trade-off between torque control bandwidth and output impedance, restricting their dynamic adaptability in complex tasks and varying scenarios\cite{2021MMT}. In contrast, the controllable stiffness of VSAs offers a promising solution to overcome this limitation\cite{2013vsa-ut}.

Due to the need for simultaneous torque transmission and stiffness variation, VSAs typically adopt a dual-motor architecture\cite{Dual-motor}. A common approach for varying stiffness involves adjusting the spring preload, which generally requires specialized components with curved or inclined surfaces for spring compression, as exemplified by MACCEPA 2.0\cite{2009maccepa}, TPS-VSA\cite{2024TPS}, and RVSA\cite{2021RVSA}. Furthermore, inspired by the human musculoskeletal system, antagonistic systems are employed for preload adjustment, utilizing opposing motors to collaboratively modulate stiffness and output torque\cite{2013Review}. Although variable stiffness methods based on adjustable preload are effective in practical applications, they have certain limitations. First, the joint cannot achieve complete rigidity. Second, when the spring is compressed or stretched, the stored potential energy cannot be directly utilized at the actuator output until the preload is released, significantly reducing the actuator's energy efficiency\cite{2023TMECH},\cite{2012AWAS}.

\par The variable lever mechanism is another approach to implementing variable stiffness. A typical lever consists of three principal points: the spring point, the pivot point, and the load point\cite{2024TMECH}. By integrating an additional motor in series to adjust the lever ratio of the load point and spring point relative to the pivot point, the spring compression resulting from the output torque can be directly controlled\cite{2018MMT}. Examples of actuators employing this design include AWAS \cite{2012AWAS}, HVSA\cite{2012HVSA}, REGT-VSA\cite{2022REGT-VSA}, and CRM-VSA\cite{2023CRM-VSA}. In this series configuration, the VSAs typically employ a stiffness motor significantly smaller than the position motor to improve compactness\cite{2011Compact-VSA}. However, a notable drawback is that the output power depends entirely on the capacity of the position motor, which limit the overall performance of the actuator. To overcome this limitation, a parallel linkage mechanism has been adopted in \cite{2011TRO}, enabling both motors to contribute simultaneously to output power. Moreover, as maintaining stiffness does not consume additional energy, the parallel configuration demonstrates superior energy efficiency\cite{2023PVSA,zhang2025design}.

However, the aforementioned VSAs commonly exhibit a significant drawback: a strong coupling between stiffness and output torque\cite{2019nonlinear}. Specifically, maintaining constant stiffness during torque output requires continuous adjustment of the stiffness setting, as the deflection angle inherently affects the characterized stiffness \cite{2016Review}. This coupling complicates the modeling and control processes, especially in applications requiring stable and precisely maintained output stiffness. In \cite{2012HVSA} and \cite{2022REGT-VSA}, the stiffness modeling process neglects the influence of the deflection angle. While this simplification might be acceptable for small deflection angles, the discrepancy between actual stiffness and model predictions increases significantly as the deflection angle grows.

To address these issues, this paper proposes a VSA with decoupled stiffness and output characteristics, referred to as the DSO-VSA, which integrates a novel Variable Stiffness Mechanism (VSM) and a Differential Transmission Mechanism (DTM). The main contributions of this paper are as follows:

\begin{enumerate}[1)]
    \item Design of a lever-based VSM that decouples stiffness from the deflection angle via a variable-length arm and employs a hypocycloidal straight-line mechanism to achieve linear torque-deflection angle behavior and wide-range continuous stiffness modulation.

    \item A DTM based on a planetary gear system is developed, which combines the power of two motors at the output according to a predefined ratio, thereby enhancing power density and output torque capability.

     \item A cascade PI controller is proposed, in which the position-loop term directly applied to stiffness and deflection angles, thereby mitigating undesired stiffness fluctuations and output disturbances.
    
\end{enumerate}

The remainder of this paper is organized as follows. Section \uppercase\expandafter{\romannumeral2} describes the mechanical design of the DSO-VSA. The design of the cascade PI controller will be introduced in Section \uppercase\expandafter{\romannumeral3}. The prototype design and the hardware setup are described in Section \uppercase\expandafter{\romannumeral4}. Experiments are conducted in Section \uppercase\expandafter{\romannumeral5} to verify the performance of the DSO-VSA. Finally, the discussion and conclusion are given in Section \uppercase\expandafter{\romannumeral6}.

\section{Mechanical Design}

\begin{figure}[t]
	\centering
	\includegraphics[width=0.9\linewidth]{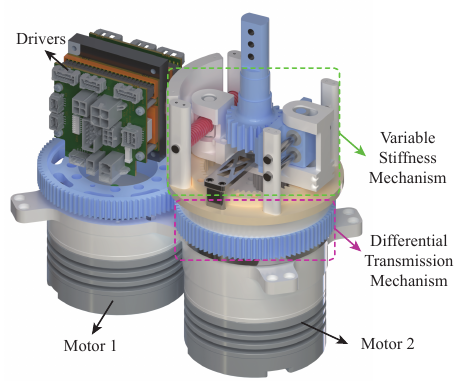}
	\caption{CAD model of the DSO-VSA.}
	\label{FIG1}
\end{figure}

The mechanical design of the DSO-VSA will be presented in this section. The main concept of the DSO-VSA is a variable stiffness actuator with decoupled stiffness and output characteristics, as shown in Fig. \ref{FIG1}. The innovative mechanical design primarily consists of the VSM and the DTM. The VSM incorporates a variable lever mechanism and a hypocycloidal straight-line mechanism. The former utilizes a variable-length arm to establish a linear torque-deflection angle relationship, thereby simplifying modeling and control, while the latter ensures a sufficient wide stiffness range. Moreover, the DTM adopts a planetary gear system, allowing the rotation of both motors to influence stiffness variation and torque output, thereby combining power from both motors at the output according to a predefined ratio.

\subsection{Design of the Variable Stiffness Mechanism}

\begin{figure}[t]
	\centering
	\includegraphics[width=\linewidth]{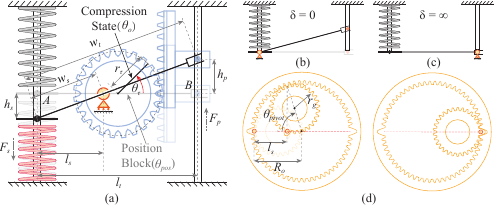}
	\caption{The principle of the VSM. (a) The analysis of the decoupled stiffness and output characteristics. (b) Zero stiffness with $l_s=0$. (c) Infinite stiffness with $l_s=l_t$. (d) The analysis of hypocycloidal straight-line mechanism. }
	\label{FIG2}
\end{figure}

The stiffness of the VSM is provided by the linear spring, and its variation results from changes in the lever ratio caused by adjusting the pivot position, as shown in Fig. \ref{FIG2}. The VSM mainly consists of a pair of spring blocks, an output block, a lever arm, and a hypocycloidal straight-line mechanism. Each spring block is connected to the left end of the lever arm at point A, while the output block is connected to the right end at point B. Based on the characteristics of the hypocycloidal straight-line mechanism, the pivot can move along the lever arm from point A to point B, enabling continuous stiffness variation from nearly zero to infinity, as shown in Fig. \ref{FIG2}(b) and \ref{FIG2}(c).

In the VSM, the characterized stiffness $\delta$ is determined by the spring stiffness $k_s$, the pivot position $l_s$, the total stroke of the pivot $l_t$, and the reference circle radius of the gear shaft $r_\tau$. In the actual system, $k_s$, $l_t$, and $r_\tau$ typically remain constant. Therefore, the characterized stiffness can be regulated by adjusting $l_s$ based on the hypocycloidal straight-line mechanism, which mainly consists of an internal gear and a pivot gear with a pivot pin, as shown in Fig. \ref{FIG2}(d). The internal gear rotates around its own center, while the pivot gear not only rotates independently but also revolves around the center of the internal gear. To ensure that the motion trajectory of the pivot pin follows the hypocycloidal straight-line characteristics, the reference circle radius of the pivot gear $r_g$ is set to half that of the internal gear $R_0$, i.e., $R_0 = 2r_g$. Therefore, $l_s$ can be expressed as follows:
\begin{equation}
l_s=R_0\left(1-\cos \theta_{pivot}\right)
\end{equation}
where $\theta_{pivot}$ is the revolution angle of the pivot gear relative to the internal gear. 

\begin{figure}[t]
	\centering
	\includegraphics[width=0.9\linewidth]{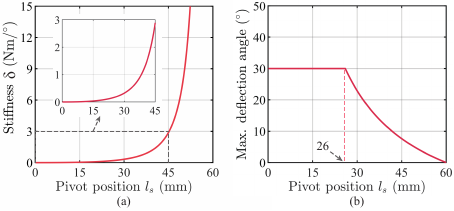}
	\caption{(a) The stiffness $\delta$ versus $l_s$. (b) The maximum deflection angle $\theta_{\tau,max}$ versus $l_s$, when $l_s<26$, $\theta_{\tau,max}$ is limited by mechanical interference, when $60>l_s>26$, $\theta_{\tau,max}$ is constrained by the maximum deflection of the spring.}
	\label{FIG3}
\end{figure}

The output block consists of a gear shaft and a modified rack. The gear shaft is optimally designed to engage with the modified rack, thereby converting rotational motion of the gear shaft into linear motion of the modified rack along the spring compression direction. Assuming that under the external load torque $\tau$, the deflection angle  between the compression state of the gear shaft and the position of the internal gear is $\theta_\tau = \theta_{o} - \theta_{pos}$, the translational distance of the modified rack $h_p$ and $\tau$ can be expressed as follows:
\begin{equation}
h_p=\theta_\tau \cdot r_\tau\\
\end{equation}
\begin{equation}
\tau=F_p \cdot r_\tau
\end{equation}
where $F_p$ is the force acting on the modified rack, and the corresponding compression length and elastic force of the spring are denoted as $h_s$ and $F_s$, respectively. Additionally, if $w_e$ and $w_t$ represent the effective lever arm length and the total lever arm length, respectively, then based on the principle of equilibrium moment at the fulcrum and the similar triangles, the following two equations can be derived:
\begin{equation}
 \frac{F_p}{F_s}=\frac{k_p h_p}{k_s h_s}=\frac{l_s}{l_t-l_s} 
\end{equation}
 \begin{equation}
 \frac{h_p}{h_s}=\frac{w_t-w_s}{w_s}=\frac{l_t-l_s}{l_s}
\end{equation}
where $k_p$ is the equivalent compression stiffness of the modified rack. Therefore, $\tau$ can be expressed as follows:
\begin{equation}
\tau=F_p r_\tau=r_\tau^2 \theta_\tau k_s\left(\frac{l_s}{l_t-l_s}\right)^2
\end{equation}
Thus, $\delta$ can be expressed as follows:
\begin{equation}
\delta=r_\tau^2 k_s\left(\frac{l_s}{l_t-l_s}\right)^2
\label{Eq7}
\end{equation}
because $\delta=\partial \tau / \partial \theta_\tau$. (7) indicates that the characterized stiffness is proportional to $r^2_\tau$ and $k_s$ but is independent of $\theta_\tau$. When $r_\tau$, $k_s$, and $l_t$ remain constant, the stiffness depends solely on $l_s$. Specifically, when $l_s=0$, the pivot position is located at point $A$, and the stiffness is zero, as shown in Fig. \ref{FIG2}(b). Conversely, when $l_s=l_t$, the pivot position shifts to point $B$, and the stiffness is nearly infinity, as shown in Fig. \ref{FIG2}(c). According to the spring selection principle proposed in \cite{2024Chris}, a pair of compression springs with a stiffness of 81.7 N/mm and a maximum deflection of 6 mm is selected. Based on these parameters, the curve of $\delta$ as a function of $l_s$ is shown in Fig. \ref{FIG3}(a).

Additionally, it is important to note that when the stiffness is very small, a large deflection angle occurs for a given load torque, which is unreasonable for robot joint constrained by mechanical interference. The relationship between the maximum deflection angle $\theta_{\tau,max}$ and $l_s$ can be expressed as follows:
\begin{equation}
\theta_{\tau, max }=\frac{h_{p, max }}{r_\tau}=h_{s, max } \frac{l_t-l_s}{r_\tau l_s}
\end{equation}
where $h_{p,max}$ is the maximum displacement of the modified rack, and $h_{s,max}$ is the maximum deflection of the spring. In the proposed VSM, due to the constraints of a compact mechanism, $\theta_{\tau.max}$ is set to 30$^\circ$, corresponding to $h_{p,max}$ = 7.85 mm for a minimum pivot position $l_s$ = 0. The relationship between $\theta_{\tau,max}$ and $l_s$ is shown in Fig. \ref{FIG3}(b). When $l_s<$ 26 mm, $\theta_{\tau,max}$ is limited by mechanical interference, whereas for 26mm $<l_s<$ 60mm, $\theta_{\tau,max}$ is constrained by the maximum deflection of the spring.

\begin{figure}[t]
	\centering
	\includegraphics[width=\linewidth]{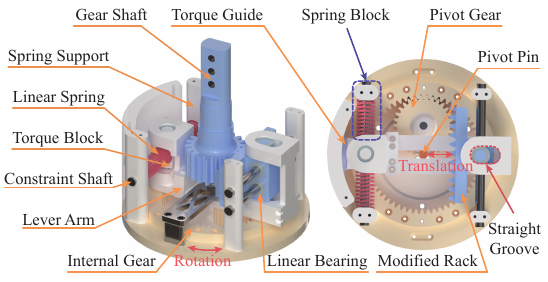}
	\caption{The mechanical composition of the VSM.}
	\label{FIG4}
\end{figure}
The mechanical composition of the VSM is shown in Fig. \ref{FIG4}. The key to achieving the characteristic of decoupling stiffness from output lies in two aspects. First, during system rotation, both the total lever arm length and the effective lever arm length must be variable. Second, the rotation joint centers of points A, B, and the pivot must be aligned with the direction of the lever arm, as shown in Fig. \ref{FIG2}(a).

The spring block consists of a linear compression spring, a spring support, and a constraint shaft. One end of the spring is fixed to the spring support, while the other end can be compressed by the torque block along the torque guide. The torque block is connected to the left end of the lever arm through a pair of bearings, allowing free rotation of the lever arm. The right end of the lever arm features a straight groove, which engages with the bearing mounted at the top and bottom of the modified rack to provide rotational and translational constraints. Additionally, a pair of linear bearings is installed at the right end of the modified rack, working with a pair of constraint shafts to maintain linear motion along the spring compression direction. Consequently, regardless of the gear shaft's rotation, the proposed VSM maintains a consistent lever ratio, thereby decouple stiffness from deflection angle.

\subsection{Design of the Differential Transmission Mechanism}

\begin{figure}[t]
	\centering
	\includegraphics[width=\linewidth]{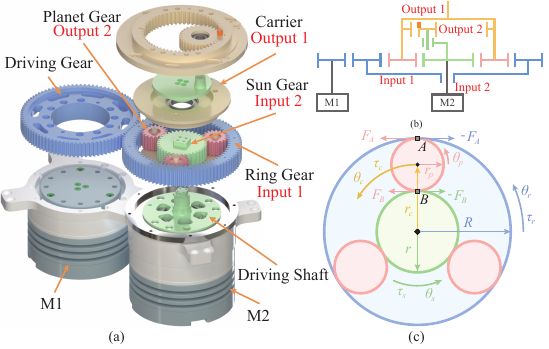}
	\caption{Basic concept of the DTM. (a) CAD model. (b) The kinematical diagram. (c) The schematic.}
	\label{FIG5}
\end{figure}

In the VSM, the stiffness and output torque are governed by two degrees of freedom (DoFs): the position of the internal gear $\theta_{pos}$ and the revolution angle of the pivot gear relative to the internal gear $\theta_{pivot}$. These two parameters correspond to the dual outputs of the DTM, which is based on a planetary gear system with differential drive. This configuration enables both motors to contribute to stiffness variation and power output, as shown in Fig. \ref{FIG5}.

The planetary gear system primarily consists of four parts: the ring gear, the sun gear, the planet gears, and the carrier. In the proposed DTM, the ring gear is designed with both internal and external teeth, where the internal teeth mesh with the planet gears, and the external teeth are driven by a driving gear mounted on the shaft of Motor 1 (M1). The sun gear, driving shaft, and pivot platform are coaxially fixed via a mortise-and-tenon structure, enabling them to rotate synchronously under the drive of Motor 2 (M2). Driven simultaneously by the ring gear and sun gear, the planet gears exhibit both self-rotation around their own axes and revolution around the sun gear. The carrier, connected to multiple planet gears via bearings, converts the planet gears' revolution into the carrier's rotation. The kinematical
diagram of the planetary gear system's transmission is shown in Fig. \ref{FIG5}(b). Therefore, the rotation radius of the planet gears $r_p$ and the carrier $r_c$ can be expressed as follows:
\begin{equation}
r_p=\frac{R-r}{2},  r_c=\frac{R+r}{2}
\end{equation}
where $R$ and $r$ are the reference circle radii of the internal teeth of the ring gear and the sun gear, respectively. As shown in Fig. \ref{FIG5}(c), since the planet gears mesh with both the ring gear and the sun gear, the rotation angles of the ring gear $\theta_r$, sun gear $\theta_s$, planet gears $\theta_p$, and carrier $\theta_c$ are not entirely independent. Based on the principle of gear transmission, which states that the arc length of the contact points of two meshing gears is equal, the following constraint can be obtained as follows:
\begin{equation}
\left\{\begin{array}{l}
\theta_r R=\theta_c r_c+\theta_p r_p \\
\theta_s r=\theta_c r_c-\theta_p r_p
\end{array}\right.
\end{equation}

Then, $\theta_p$ and $\theta_c$ can be expressed as follows:
\begin{equation}
\theta_p=\frac{\theta_r R-\theta_s r}{2 r_p}=\frac{\theta_r R-\theta_s r}{R-r} 
\end{equation}
\begin{equation}
\theta_c=\frac{\theta_r R+\theta_s r}{2 r_c}=\frac{\theta_r R+\theta_s r}{R+r}
\label{Eq12}
\end{equation}

(\ref{Eq12}) indicates that the rotation angle of the carrier is a weighted average of the rotation angles of the ring gear and the sun gear, with weights proportional to their respective reference circle radii. 
%This implies that both the ring gear and the sun gear tend to influence the carrier's motion in proportion to their respective reference pitch radii.
Additionally, it should be noted that the coordinate system of $\theta_p$ is absolute. Therefore, when taking the sun gear as the reference coordinate origin, the relative rotation angle of the planet gears $\theta_{c,s}$ can be expressed as follows:
\begin{equation}
\theta_{c, s}=\theta_c-\theta_s=\frac{R \left(\theta_r-\theta_s\right)}{R+r}
\end{equation}

\begin{figure*}[t]
	\centering
	\includegraphics[width=\textwidth]{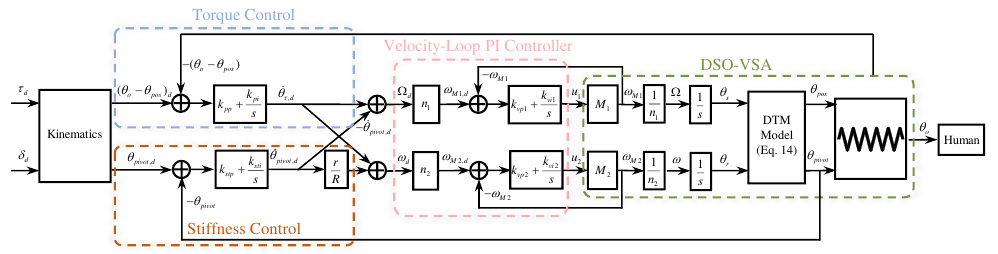}
	\caption{Block-diagram of the proposed cascade PI controller.}
	\label{FIG6}
\end{figure*}

Therefore, when the rotations of the ring gear and sun gear serve as Input 1 and Input 2 of the DTM, the two outputs of the DTM-corresponding to the two DoFs of the VSM-are linearly coupled, as follows:
\begin{equation}
\left[\begin{array}{c}
\theta_{pos} \\
\theta_{pivot}
\end{array}\right]=
\left[\begin{array}{c}
\theta_c \\
\theta_{c,s}
\end{array}\right]=\left[\begin{array}{cc}
\alpha & 1-\alpha \\
\alpha & -\alpha
\end{array}\right]\left[\begin{array}{c}
\theta_r \\
\theta_s
\end{array}\right]
\label{Eq14}
\end{equation}
where
\begin{equation}
\alpha=\frac{R}{R+r}
\end{equation}

Similarly, the torque distribution of the planetary gear system can be derived. Assuming the input torques of the ring gear and the sun gear are $\tau_r$ and $\tau_s$, respectively. When the system is in equilibrium, the following constraint can be obtained as follows:
\begin{equation}
\left\{\begin{array}{l}
\tau_r-F_B R=0 \\
\tau_s-F_A r=0
\end{array}\right.
\end{equation}
where $F_A$ and $F_B$ are the forces exerted on the planet gears by the ring gear and the sun gear, respectively. Assuming the output torque of the carrier is $\tau_c$, since the rotation of the planet gears and the carrier are in equilibrium, the condition $F_A=F_B$ must hold, and the sum of the external forces acting on the carrier must be zero, as follows:
\begin{equation}
\frac{\tau_c}{r_c}+F_A+F_B=0
\end{equation}

Therefore, the torques acting on the ring gear, sun gear, and carrier satisfy the proportional relationship as follows:
\begin{equation}
\tau_r: \tau_s: \tau_c=-R:-r:(R+r)
\end{equation}

Since the rotation of the carrier follows (\ref{Eq12}), the power distribution of the DTM is as follows:
\begin{equation}
P_r: P_s: P_c=\Omega R: \omega r:(\Omega R+\omega r)
	\label{Eq19}
\end{equation}
where $P_r$, $P_s$, and $P_c$ are the output power of the ring gear, the sun gear, and the carrier, respectively. $\Omega$ and $\omega$ are the rotation velocities of the ring gear and the sun gear, respectively. (\ref{Eq19}) indicates that the output power of the carrier is distributed between the ring gear and sun gear in a designed ratio. This ratio depends not only on the reference circle radii of the ring gear and sun gear but also on their rotation velocities. Therefore, based on the planetary gear system, the DTM enables both motors to jointly contribute to power output, endowing the DSO-VSA with dual-motor load sharing capability.

\section{Torque Controller Design}

When the DSO-VSA is in the compliant state ($l_s <$ 60 mm), a modified cascade PI controller incorporating feedback from each motor is designed. Unlike traditional cascade PI controllers that directly generate control signals based on the feedback stiffness and torque signals, the modified cascade PI controller proposed in this paper utilizes the established kinematic model to equivalently transform stiffness and torque regulation into position control tasks, as shown in Fig. \ref{FIG6}.

Specifically, based on (\ref{Eq14}), the desired stiffness angle $\theta_{pivot,d}$ and the desired deflection angle $\theta_{\tau,d}=(\theta_0-\theta_{pos})_d$ can be calculated according to the desired stiffness $\delta_d$ and the desired output torque $\tau_d$. The cascade PI controller is to control the rotation of the two motors based on the established kinematic model to track $\theta_{pivot,d}$ and $\theta_{\tau,d}$. However, the set point control method introduced in \cite{2012HVSA} lacks consideration of velocity variables during rotation, which may lead to undesired stiffness variations and torque fluctuations, posing potential safety risks. Therefore, the proposed controller directly applies the position-loop control term to stiffness angle and deflection angle regulation processes, as shown below:
\begin{equation}
		 \dot{\theta}_{pivot, d}=\left(k_{s t p}+\frac{k_{s t i}}{s}\right)\left(\theta_{pivot, d}-\theta_{pivot}\right) 
\end{equation} 
\begin{equation}		 
		 \dot{\theta}_{\tau,d}=\left(k_{p p}+\frac{k_{p i}}{s}\right)\left(\theta_{\tau, d} - \theta_\tau\right)
\end{equation}
where $s$ is the Laplace operator. $\dot\theta_{pivot,d}$ and $\dot\theta_{\tau,d}$ are the desired stiffness variation velocity and desired rotation velocity, respectively. In the DTM, the synchronous rotational component of the ring gear and the sun gear generates the output torque, while their differential component causes stiffness variation. Therefore, the two motors must proportionally allocate the rotation velocity adjustments induced by torque generation and stiffness variation. The specific relationship is as follows:
\begin{equation}
		\omega_{M1,d}=n_1 \Omega_d=n_1\left(\dot{\theta}_{\tau,d}- \dot{\theta}_{pivot,d }\right) 
\end{equation}		
\begin{equation}
		\omega_{M2,d}=n_2 \omega_d=n_2\left(\dot{\theta}_{\tau,d}+\frac{r}{R} \dot{\theta}_{pivot, d}\right)
\end{equation}
where $n_1$ and $n_2$ are the transmission ratios from the ring gear to M1 and from the sun gear to M2, respectively. The velocity-loop PI controllers for the two motors are as follows:
\begin{equation}
		 u_1=\left(k_{v p 1}+\frac{k_{v i1}}{s}\right)\left(\omega_{M1, d}-\omega_{M1}\right) 
\end{equation}		
\begin{equation}
		 u_2=\left(k_{v p 2}+\frac{k_{v i 2}}{s}\right)\left(\omega_{M2, d}-\omega_{M2}\right)
\end{equation}
where $\omega_{M1,d}$ and $\omega_{M1}$, as well as $\omega_{M2,d}$ and $\omega_{M2}$, are the desired and actual rotation velocities of M1 and M2, respectively. Then, the control laws for the two motors can be expressed as follows:
\begin{equation}
	\begin{aligned}
	u_1=\left(k_{v p 1}+\frac{k_{v i 1}}{s}\right ) \left\{n_1 \left[\left (k_{p p}+\frac{k_{p i}}{s}\right ) (\theta_{\tau,d}-\theta_\tau) \right. \right.
	\\ \left.\left.- \left(k_{s t p}+\frac{k_{s t i}}{s}\right) (\theta_{pivot, d}-\theta_{pivot})\right]-\omega_{M1}\right\}
	\end{aligned}
\end{equation}
\begin{equation}
	\begin{aligned}
	u_2=\left(k_{v p 2}+\frac{k_{v i 2}}{s}\right )\left\{n_2\left[\left (k_{p p}+\frac{k_{p i}}{s}\right)(\theta_{\tau,d}-\theta_\tau)\right. \right.
	\\ \left.\left.+\frac{r}{R} \left(k_{s t p}+\frac{k_{s t i}}{s}\right)\left(\theta_{pivot, d}-\theta_{pivot}\right)\right]-\omega_{M2}\right\}
	\end{aligned}
\end{equation}

\section{Prototype Design}

The prototype of the DSO-VSA, which features decoupled stiffness and output characteristics, is shown in Fig. \ref{FIG7}(a). This prototype consists of a VSM and a DTM. Within the VSM, stiffness is provided by a pair of rectangular springs, each possessing a stiffness coefficient of 81.7 N/mm. Due to the constraints of a compact mechanism design, the internal gear's reference circle radius is set at 30 mm, which also defines the maximum stroke of the pivot. In the DTM, the reference circle radii of the ring gear and sun gear have a ratio of 2:1 (36 mm : 18 mm), indicating that under constant stiffness conditions, their contributions to the output power also follow this 2:1 proportion.

To provide sufficient torque, the total transmission ratio between the 150W M1 (EC Frameless DT50-M, Maxon Motor AG, Sachseln, Switzerland) and the ring gear is set to -100:1, whereas the transmission ratio between the 80W M2 (EC60 Flat, Maxon Motor AG, Sachseln, Switzerland) and the sun gear is 50:1. Driven by these two motors, the DSO-VSA delivers a nominal torque of 55.5 Nm and a nominal rotation velocity of 43 RPM. Two encoders, each with a resolution of 8.8$\times$10$^{-2}$ $^\circ$/pulse, measure the positions of the driving gear and the driving shaft. All characteristics of the DSO-VSA are summarized in Table \uppercase\expandafter{\romannumeral1}.

\begin{table}[h]
	\caption{\textbf{DSO-VSA Characteristics}}%æ é¢
	\centering%æè¡¨å±ä¸­
	\begin{tabular*}{\hsize}{@{}@{\extracolsep{\fill}}lll@{}}%åä¸ªcä»£è¡¨è¯¥è¡¨ä¸å±ååï¼åå®¹å¨é¨å±ä¸­
		\toprule%ç¬¬ä¸éæ¨ªçº¿
		\textbf{Parameter}&\textbf{Value}&\textbf{Unit} \\
		\midrule%ç¬¬äºéæ¨ªçº¿ 
		Nominal torque&55.7&N$\cdot$m \\
		Range of motion ($\theta_{pos}$)&$\pm$180&deg \\
		Spring stiffness ($k$)&81.7&kN/m \\
		Range of stiffness ($\delta$)&(0, $\infty$)&Nm/deg \\
		Max. deflection angle&30&deg \\
		Nominal rotation velocity&43&rpm \\
		Stiffness variation time&0.82&s \\
		Max. elastic energy&1.47&J \\
		Weight (without motors)&1.24&kg \\
		\bottomrule%ç¬¬ä¸éæ¨ªçº¿
	\end{tabular*}
\end{table}

\section{Experiments and Results}

\begin{figure}[t]
	\centering
	\includegraphics[width=\linewidth]{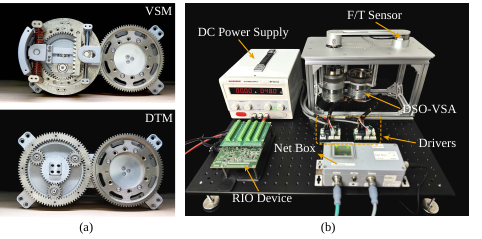}
	\caption{The prototype of the DSO-VSA and the experimental setup.}
	\label{FIG7}
\end{figure}

To evaluate the performance of the DSO-VSA, an experimental platform was established, as shown in Fig. \ref{FIG7}(b). The platform consists of a DSO-VSA prototype, an RIO device, a DC power supply, an output link, a force/torque sensor and two drivers. The experiments include stiffness calibration, stiffness regulation, torque tracking and step response, and evaluation of the decoupled characteristics and the dual-motor load sharing capability. Five trials were done for each setup.

\subsection{Stiffness Calibration}

The pivot position of the VSM can be regulated between 0 mm and 60 mm, corresponding to a theoretical stiffness range from 0 to nearly infinity. In the stiffness calibration experiment, the relationship between output torque and deflection angle was measured, with the pivot position adjustment step set to 5 mm within the defined boundary conditions. Since the maximum deflection angle described in Section \uppercase\expandafter{\romannumeral2}, is limited by the mechanical interference or the maximum deflection of the spring, the desired deflection angle under each pivot position step is set to 80$\%$ of the maximum deflection angle.

\begin{figure}[h]
	\centering
	\includegraphics[width=\linewidth]{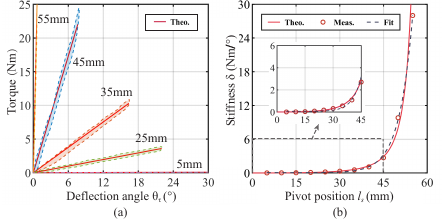}
	\caption{Stiffness calibration experiments. (a) Torque-deflection angle curves of the DSO-VSA across 5 selected stiffness values. (b) Characterized stiffness varies with the change in pivot position.}
	\label{FIG8}
\end{figure}

For clarity, 5 out of 11 experimental curves were selected for display, as shown in Fig. \ref{FIG8}(a). The dashed lines represent experimental data, while the solid lines indicate theoretical results. The experimental results show that the output torque exhibits an approximately linear relationship with the deflection angle, primarily due to the decoupling characteristics between stiffness and output torque. Additionally, when the torque variation trend reverses (e.g., from increasing to decreasing), slight hysteresis and deviations appear in the curves. This phenomenon is likely caused by component frictions, while machining errors and gear backlash may also contribute to it.

Fig. \ref{FIG8}(b) show the stiffness variation at different pivot positions. The red circles represent stiffness values calculated from the slope of the torque-deflection angle curve for each fitted dataset. The experimental curve is fitted while the theoretical curve is derived from (\ref{Eq7}). The close correlation between experimental results and theoretical predictions verifies the accuracy of the model. Therefore, in future applications with different spring configurations, the stiffness curve of the DSO-VSA can be directly predicted based on the theoretical model without the need for experimental validation.

\subsection{Stiffness Regulation}

\begin{figure}[t]
	\centering
	\includegraphics[width=0.9\linewidth]{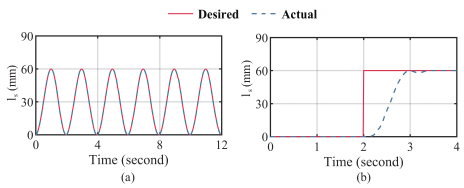}
	\caption{The stiffness regulation experiment, where $l_s$ is selected as the stiffness characteristic index, with the movement of $l_s$ from 0 to 60 mm corresponding to a stiffness variation ranging from 0 to infinity. (a) Sinusoidal signal tracking. (b) Step response.}
	\label{FIG9}
\end{figure}

The stiffness of the DSO-VSA is determined by the pivot position. To clearly demonstrate the capability of the DSO-VSA to vary stiffness across its entire operational range, $l_s$ is selected as the stiffness characteristic index, where the movement of $l_s$ within the range of 0 mm to 60 mm corresponds to a stiffness variation from 0 to infinity.  Fig. \ref{FIG9}(a) and \ref{FIG9}(b) show the sinusoidal signal tracking and step response of $l_s$, respectively. The amplitude of the sinusoidal signal is 30 mm with a frequency of 0.5 Hz. Meanwhile, the input step of the step response is 60 mm, representing the adjustment range from the minimum stiffness to the maximum stiffness. In the sinusoidal signal tracking experiment, the root mean square (RMS) error of $l_s$ is 0.09 mm. In the step response experiment, the 90$\%$ rise time is 0.824 $\pm$ 0.028 s (mean $\pm$ s.d.). In addition, after the step response rises to 60 mm, a slight drop in pivot position followed by a rapid return to the set value is observed. This phenomenon is attributed to the motion characteristics of the hypocycloidal straight-line mechanism, which induce a reversal in the overshoot direction of the pivot position.

\subsection{Torque Control Performance}

\begin{figure}[h]
	\centering
	\includegraphics[width=0.95\linewidth]{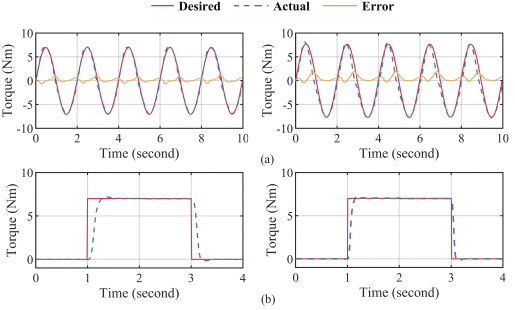}
	\caption{Experimental results of the torque control performance under low stiffness (left) and high stiffness (right). (a) Sinusoidal torque tracking with frequency of 0.5 Hz and amplitudes of 7 Nm. (b) Step response with step input of 7 Nm.}
	\label{FIG10}
\end{figure}

The torque control performance of the DSO-VSA equipped with the modified cascade PI controller is evaluated through the sinusoidal torque tracking and the step response, as shown in Fig. \ref{FIG10}. To explore the effect of different stiffness parameters on torque control performance, two stiffness levels are set as experimental conditions: low stiffness ($l_s$ = 30 mm, $\delta$ = 0.321 Nm/$^\circ$) and high stiffness ($l_s$ = 45 mm, $\delta$ = 2.888 Nm/$^\circ$). Under these two stiffness conditions, sinusoidal torque tracking and step response are conducted, respectively. In the torque tracking experiment, a sinusoidal torque signal with an amplitude of 7 Nm and a frequency of 0.5 Hz is applied, while in the step response experiment, the step input is set as 7 Nm.

The results of torque tracking experiment are shown in Fig. \ref{FIG10}(a). Under the low stiffness condition, the RMS error of the sinusoidal torque tracking is 0.32 Nm, whereas under the high stiffness condition, this error increases to 0.89 Nm. This significant increase primarily due to the elevated stiffness amplifies the torque error caused by the system response time and assembly clearance. Fig. \ref{FIG10}(b) shows the results of the step response. Compared with the low stiffness condition, where the 90$\%$ rise and fall times are 0.208 $\pm$ 0.012 s and 0.154 $\pm$ 0.016 s, respectively, these times are reduced to 0.116 $\pm$ 0.008 s and 0.078 $\pm$ 0.011 s under the high stiffness condition. This decreased trend is mainly attributed to the reduced deflection angle required to reach the same desired torque with increased stiffness, thereby shortening the necessary motion time. Additionally, the fall time is significantly shorter than the rise time, which may be attributed to the system needing to overcome the load torque during the step-up process.
%whereas during the step-down process, the elastic energy stored in the spring is released, accelerating the response speed.

% ä½ååº¦ä¸çæ­£å¼¦æ­ç©è·è¸ªçRMSè¯¯å·®æ¯0.12Nmï¼èå¨é«ååº¦ä¸åå¢å å°äº0.46Nmï¼è¿ç§è¯¯å·®å¢å¤§çè¶å¿ä¸»è¦å½å äºæ´é«çååº¦æ¾å¤§äºç±äºç³»ç»ååºæ¶é´åå®è£é´éæå¯¼è´çæ­ç©è¯¯å·®ãå¾10(b)æ¯é¶è·ååºå®éªçå®éªç»æï¼å¨ä½ååº¦åé«ååº¦ä¸ï¼90%çä¸åæ¶é´åå«æ¯0.8så0.4sï¼èå¯¹åºçä¸éæ¶é´åå«æ¯0.6så0.2sã
%å¨ä½ååº¦ä¸ï¼90%çä¸ååä¸éæ¶é´ï¼the 90% rise and fall timesï¼åå«æ¯0.8så0.6sï¼èå¨é«ååº¦ä¸ï¼90%çä¸ååä¸éæ¶é´åå«æ¯0.4så0.2sãè¿ç§åå°çè¶å¿å½å äºå¨å°è¾¾ç¸åçæææ­ç©æ¶ï¼æ´é«çååº¦å¯¹åºæ´å°çåè½¬è§åº¦ï¼è¿ç¼©ç­äºè¿å¨æ¶é´ãæ­¤å¤ï¼Compared to the rise time, the fall time is significantly shorter, which may be attributed to the system needing to overcome torque work during the step-up process, whereas during the step-down process, the elastic energy stored in the spring is released, accelerating the response speed.

%30 å¯¹åº0.32
%45 å¯¹åº2.89

%ï¼  ç¶èï¼æ´é«çååº¦çé¶è·ä¸åæ¶é´ç­ãåå å¨äºæ´é«çååº¦ç¼©ç­äºå°è¾¾ç¸åæ­ç©çæ¶é´

\subsection{Decoupled characteristics}

\begin{figure}[h]
	\centering
	\includegraphics[width=\linewidth]{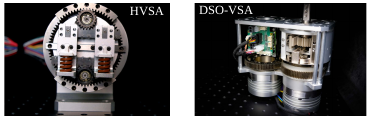}
	\caption{The prototype of the HVSA (left) and the DSO-VSA (right).}
	\label{FIG11}
\end{figure}

\begin{figure}[h]
	\centering
	\includegraphics[width=\linewidth]{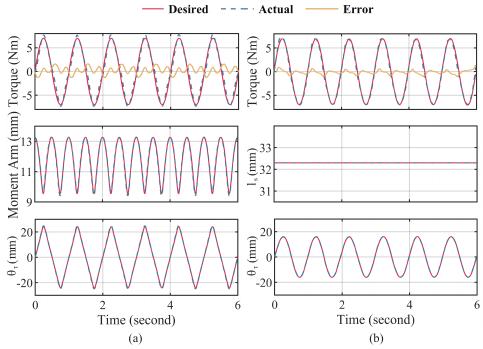}
	\caption{Experimental validation of decoupled characteristics, where the moment arm (pivot position) and deflection angle during output are compared between (a) HVSA and (b) DSO-VSA.}
	\label{FIG12}
\end{figure}

To validate the advantages of the decoupled stiffness and output characteristics of the DSO-VSA, we reproduced the HVSA prototype described in \cite{2012HVSA} and conducted comparative experiments, as shown in Fig. \ref{FIG11}. In contrast to the original study, the modeling process for the reproduced HVSA in this work did not neglect the influence of the deflection angle on stiffness. Both prototypes were controlled under the same controller described in Section \uppercase\expandafter{\romannumeral3}, with constant stiffness, and tasked with tracking a sinusoidal torque signal of 7 Nm amplitude and 1 Hz frequency. During the experiment, motor positions were recorded in real time to calculate the moment arm (pivot position) and deflection angle, as shown in Fig. \ref{FIG12}.

Fig. \ref{FIG12}(a) shows the experimental results of the HVSA. Due to the inherent coupling between stiffness and output, the deflection angle variations compel the moment arm to adjust in order to maintain constant stiffness, resulting in frequent torque fluctuations. Meanwhile, dynamic variations in the moment arm significantly influence the deflection angle, causing it to exhibit an approximately triangular waveform. In contrast, the DSO-VSA, which features decoupled stiffness and output characteristics, exhibits smoother and reduced torque fluctuations due to its constant pivot position. These results further confirm that the decoupled characteristics enhances the stability of torque control.

\subsection{Dual-Motor Load Sharing}

% For a VSA whose output relies solely on the position motor, its output power is typically significantly lower than that of the position motor due to transmission efficiency and mechanical friction, which is unavoidable. However, due to the dual-motor load sharing capability provided by the DTM, the DSO-VSA can significantly increase its output power. To verify this capability, the dual-motor load sharing experiment required the DSO-VSA to rotate within the maximum deflection angle range at an extremely low velocity while maintaining constant stiffness.

\begin{figure}[h]
	\centering
	\includegraphics[width=7cm]{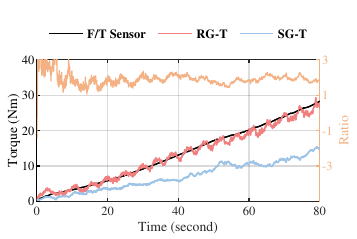}
	\caption{Verification of the dual-motor load sharing capability, where RG-T and SG-T denote the calculated torques of the ring gear and sun gear, respectively. The torque ratio is defined from RG-T to SG-T.}
	\label{FIG13}
\end{figure}

For a VSA whose output depends exclusively on the position motor, the nominal output power is typically much lower than the motor's power due to inherent transmission efficiency losses. To verify the dual-motor load sharing capability of the DSO-VSA, experiment was conducted in which the DSO-VSA rotated within its maximum deflection angle range at an extremely low velocity while maintaining constant stiffness. Throughout the experiment, motor currents were continuously monitored. The corresponding input torques for the ring gear (RG-T) and sun gear (SG-T) were calculated by applying the torque constants and accounting for the total transmission ratio. Fig. \ref{FIG13} shows the data of RG-T, SG-T, and the torque measured by the F/T sensor, along with the ratio of RG-T to SG-T. The experiment results shows that as the output torque increased, both RG-T and SG-T increased proportionally, which is consistent with theoretical expectations. Notably, the sum of RG-T and SG-T was significantly higher than the measured output torque (F/T Sensor). This discrepancy can be primarily attributed to two main factors: first, mechanical friction caused the motors to bear additional torque loads. Second, part of the motor current was consumed to overcome internal motor energy losses, including copper loss, iron loss, and other inefficiencies.

\begin{table*}[t]
	\caption{\textbf{Comparisons of different VSAs}}%æ é¢
	\centering%æè¡¨å±ä¸­
\begin{threeparttable}
	\begin{tabular*}{\textwidth}{@{}@{\extracolsep{\fill}}lllllllll@{}}%åä¸ªcä»£è¡¨è¯¥è¡¨ä¸å±ååï¼åå®¹å¨é¨å±ä¸­
		\toprule%ç¬¬ä¸éæ¨ªçº¿
		Actuator&RS(Nm/$^\circ$)&PTR(\%)&RM($^\circ$)&RDA($^\circ$)&NT(Nm)&SVT(s)&MES(J)&DSO \\
		\midrule%ç¬¬äºéæ¨ªçº¿ 
	
	SVSA\cite{2018MMT}&0.030-2.628&83&$\pm$180&$\pm$45&9.46&-&3.7&No\\
	HVSA \cite{2012HVSA}&0.070-2.200&50&$\pm$45&$\pm$30&8.5&0.16&0.86&No  \\
	REGT-VSA \cite{2022REGT-VSA}&0.349-41.225&78.9&$\pm$180&$\pm$10.7&22&0.5&4.1&No   \\
	AwAS\cite{2012AWAS}&0.524-22.689&-&$\pm$120&$\pm$14&10.75&2.5&3.5 &No             \\
	AwAS-\uppercase\expandafter{\romannumeral2}\cite{2012AWAS2}&0-$\infty$&-&$\pm$150&$\pm$17&10.75&3.5&5.8&No \\
	vsaMGR\cite{VSAMGR}&1.414-10.594&93&$\pm$180&$\pm$3.5&15&1.1&1.1&No   \\
	vsaUT-\uppercase\expandafter{\romannumeral2} \cite{VSA-UT}&0.7-948&88&$\pm$28.6&$\pm$40.1&15&0.78&0.19&No  \\
	\textbf{DSO-VSA}&\textbf{0-$\infty$}&\textbf{97.8}&\textbf{$\pm$180}&\textbf{$\pm$30}&\textbf{55.7}&\textbf{0.82}&\textbf{1.47}&\textbf{Yes}    \\
	
		\bottomrule%ç¬¬ä¸éæ¨ªçº¿
	\end{tabular*}

	\begin{tablenotes}[flushleft]
		\item Note: RS: Range of stiffness, PTR: Power transmission ratio, RM: Range of motion, RDA: Range of deflection angle, NT: Nominal torque, SVT: Stiffness variation time, MEE: Maximum elastic energy, DSO: Decoupled stiffness and output.
	\end{tablenotes}
\end{threeparttable}
\end{table*}
\section{Discussion and Conclusion}

In this paper, a variable stiffness actuator with decoupled stiffness and output characteristics, called DSO-VSA, is proposed for rehabilitation robotics. The mechanism of the DSO-VSA mainly consists of a VSM and a DTM. The proposed VSM utilizes a linear cycloidal mechanism, allowing continuous stiffness variation from zero to nearly infinity by adjusting the pivot position. Crucially, the VSM fully decouples stiffness from output characteristics, significantly enhancing stiffness control and torque output performance.  Meanwhile, the DTM, employing a planetary gear system, provides the DSO-VSA with dual-motor load sharing capabilities, thereby overcoming the limitations inherent in single-motor configurations. DSO-VSA is guided by a series of mechanical objectives, as shown in Table \uppercase\expandafter{\romannumeral1}.

To further validate the capabilities of the proposed prototype, a comparison between the DSO-VSA and several well-known VSAs is given in Table \uppercase\expandafter{\romannumeral2}. Similar to AWAS-\uppercase\expandafter{\romannumeral2} \cite{2012AWAS2}, the DSO-VSA provides stiffness range from zero to infinity. However, unlike AWAS, the DSO-VSA achieves decoupling between stiffness and deflection angle by configuring a variable total length of the lever arm while maintaining a constant rack motion direction. This decoupling characteristic results in a linear torque-deflection angle relationship, significantly simplifying system modeling and control complexity. It is worth noting that the stiffness-pivot position curve can be easily modified by adjusting certain design parameters (e.g., spring material and elastic coefficient, pivot' total stroke, etc.) to accommodate different application scenarios.

An essential motivation behind designing this VSA is to facilitate easier driving of rehabilitation exoskeleton joints. Accordingly, the nominal torque of the DSO-VSA is 55.7 Nm, and the nominal velocity can reach up to 258 deg/s. Since the nominal torque can be increased by using a larger ratio reducer, even when considering the need for nominal velocity, motors with greater power ratings can be utilized. To effectively evaluate motor efficiency in terms of their contributions to output power, we define the Power Transmission Ratio (PTR) as the ratio between the nominal actuator output power under constant stiffness conditions and the combined nominal output power of both motors. All VSAs are compared using this parameter. Benefiting from the dual-motor load sharing capability enabled by the DTM, the DSO-VSA achieves a PTR of 97.8$\%$. This indicates that, when the product of nominal torque and nominal velocity is used as a performance metric, the DSO-VSA exhibits a 96$\%$ improvement over the HVSA under identical total motor power conditions. Moreover, this performance advantage becomes even more pronounced when the HVSA employs a larger motor dedicated to stiffness variation.

Moreover, since the rotation of each motor simultaneously affects both stiffness variation and torque output, traditional set position control method lack dynamic consideration of state variables during the rotation process. To address this limitation, we designed a cascade PI controller wherein the position-loop control term directly regulates both the stiffness and output torque. Subsequently, a series of experiments were conducted to validate the actuator's stiffness calibration, stiffness regulation, torque control performance, and dual-motor load sharing capability.

Despite the promising results, the proposed DSO-VSA has several limitations. Firstly, due to the trade-off between achievable output torque and the mechanical strength required of internal components, the overall weight of the DSO-VSA is relatively high. Secondly, positioning the pivot point closer to the load contact point significantly increases the stiffness variation rate but leads to a notable reduction in resolution, mainly due to the motion characteristics associated with the variable lever ratio. Although this issue can be partially mitigated by increasing the pivot's total stroke or selecting a spring with alternative stiffness properties to shift the desired stiffness into a higher-precision region, extending the pivot stroke substantially prolongs the stiffness adjustment time. Additionally, the designed DSO-VSA prototype inherently demonstrates low-velocity, high-torque output characteristics, which is suitable for rehabilitation applications. However, for scenarios demanding extensive stiffness adjustments and rapid position tracking, this limited velocity restricts the actuator's performance. This limitation can be effectively addressed by reducing the gear ratio or utilizing a higher-power motor. Consequently, The core contribution of this work lies in proposing a unified solution that achieves stiffness-output decoupling and enables generalized dual-motor load sharing via a planetary gear mechanism.

\par In future work, we will focus on parameter optimization and modular design to further enhance the DSO-VSA's overall performance. Additionally, we plan to investigate its integration into exoskeleton systems for shoulder and elbow rehabilitation.

% References

\bibliographystyle{IEEEtran}
\bibliography{Reference}\ %IEEEabrv instead of IEEEfull

\vfill

%\vspace{-1cm}

\end{document}